%% file: acl_latex.tex
\pdfoutput=1

\documentclass[11pt]{article}

\usepackage[]{acl}

\usepackage{multicol}

\usepackage{times}
\usepackage{latexsym}
\usepackage{tcolorbox}

\usepackage[T1]{fontenc}

\usepackage{array}
\newcolumntype{P}[1]{>{\centering\arraybackslash}p{#1}}

\usepackage[utf8]{inputenc}

\usepackage{microtype}

\usepackage{bbm}
\usepackage{amsmath}
\usepackage{amssymb}

\usepackage{amsmath,amssymb,amsfonts}
\usepackage{bm}

\usepackage{xcolor,colortbl}
\usepackage{multirow}
\usepackage{booktabs}

\usepackage{graphicx}
\usepackage{subcaption}

\newcolumntype{P}[1]{>{\centering\arraybackslash}p{#1}}

\title{Assessing Distractors in Multiple-Choice Tests}

\author{Vatsal Raina  \\
  ALTA Institute, Cambridge University \\
  \texttt{vr311@cam.ac.uk} \\\And
Adian Liusie \\
  ALTA Institute, Cambridge University \\
  \texttt{al826@cam.ac.uk} \\\AND
  Mark Gales \\
  ALTA Institute, Cambridge University \\
  \texttt{mjfg@cam.ac.uk} \\}

\begin{document}
 \maketitle
\begin{abstract}

Multiple-choice tests are a common approach for assessing candidates' comprehension skills. Standard multiple-choice reading comprehension exams require candidates to select the correct answer option from a discrete set based on a question in relation to a contextual passage. For appropriate assessment, the distractor answer options must by definition be incorrect but plausible and diverse. However, generating good quality distractors satisfying these criteria is a challenging task for content creators. We propose automated assessment metrics for the quality of distractors in multiple-choice reading comprehension tests. Specifically, we define quality in terms of the incorrectness, plausibility and diversity of the distractor options. We assess incorrectness using the classification ability of a binary multiple-choice reading comprehension system. Plausibility is assessed by considering the distractor confidence - the probability mass associated with the distractor options for a standard multi-class multiple-choice reading comprehension system. Diversity is assessed by pairwise comparison of an embedding-based equivalence metric between the distractors of a question. To further validate the plausibility metric we compare against candidate distributions over multiple-choice questions and agreement with a ChatGPT model's interpretation of distractor plausibility and diversity.

\end{abstract}

\input{introduction}

\input{background}

\input{experiments}

\input{conclusions}


\bibliography{anthology,custom}
\bibliographystyle{acl_natbib}

\appendix
\input{appendix}

\end{document}

%% file: introduction.tex
\section{Introduction}
\begin{figure}[t!]
    \centering
    \includegraphics[width=1.0\columnwidth]{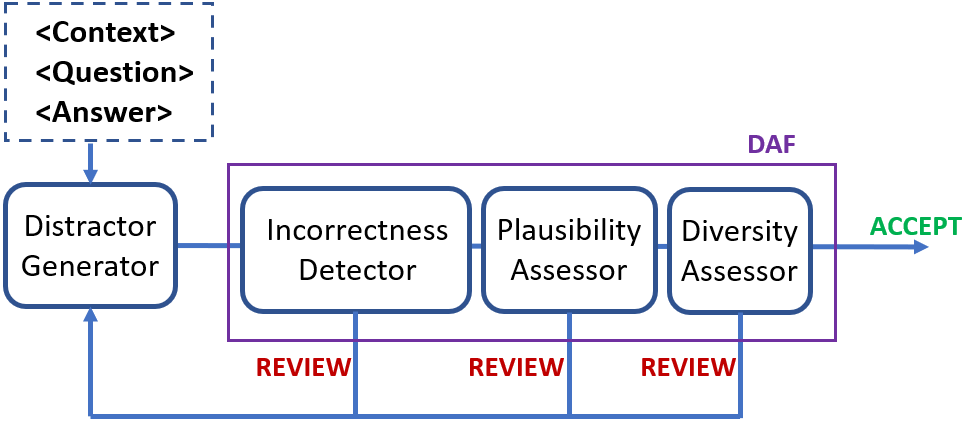}
    \caption{Distractor Assessment Framework (DAF) filtration pipeline for generated distractors.}
    \label{fig:daf_pipeline}
\end{figure}

Multiple-choice tests are an efficient and effective way of assessing candidates' comprehension skills \citep{AldersonJ.Charles.2000AR} with key advantages such as being a standardized format, eliminating subjective grading and being easy to grade. These advantages make them a highly popular assessment method widely adopted in a range of settings \citep{kurz1999review}, such as university exams, job screening and qualification accreditation. A challenging aspect of generating suitable multiple-choice questions is in selecting the incorrect options, i.e the distractors \citep{gierl2017developing}. Selecting good distractors is a subtle process, which requires the option to possess several properties \citep{qiu2020automatic}; 1) The distractor should not be a possible correct answer, as this would make marking the question subjective. 2) the distractor option should not be too obviously invalid, as then candidates may easily avoid them. 3) The questions should have relatively diverse distractors, as this would better allow questions to gauge more information from candidates. 

Currently, test creators conduct a pre-test phase where questions are internally reviewed and then tested on a subset of real candidates \citep{liusie2023analysis}, an evaluation process that is very manual and can be both subjective and expensive. Automating the process to evaluate distractors would lead to improved efficiency in the test creation process, and may aid test designers to create high-quality questions. However, currently, assessing the quality of distractors is a challenging task. To the best of our knowledge, there are no existing datasets targeted towards assisting automated distractor evaluation (beyond sequence overlap measures \citep{gao2019generating}), and therefore any approach has to port information from other resources. Further, validating the efficacy of approaches is a challenging task, especially without manual labels of distractor quality, which themselves due to the nature of the task are at risk of being subjective. 

In this paper, we propose the Distractor assessment framework (DAF), a collection of systems that can be used to automatically determine the quality of distractors. Our framework provides automatic scores for the 3 previously mentioned important properties of the distractors: incorrectness, plausibility and diversity. The incorrectness detector is a binary machine reading comprehension system that predicts whether a given distractor could be the correct answer, the plausibility evaluator leverages system confidence, while the diversity assessor considers the average similarity score between all pairs of distractors. We further propose several methods to probe existing large-scale foundation models, specifically ChatGPT instruction fine-tuned \citep{ouyang2022training} from GPT-3 \citep{brown2020language}, to validate the suitability of our quality metrics and demonstrate that our methods do reasonably capture elements of the considered properties. Additionally, we validate the plausibility metric against human candidate distributions on multiple-choice questions.
Our contributions can be summarized as follows:
\begin{itemize}
    \item Proposed assessment metrics for the challenging task of distractor assessment in terms of incorrectness, plausibility and diversity.
    \item Verification of the assessment metrics including probing ChatGPT and comparison with real candidate distribution scores.
\end{itemize}

\section{Related Work}
\label{sec:related}

Previous automatic distractor assessment methods proposed to compare the similarity of generated distractors with the ground-truth distractors present in the dataset \citep{gao2019generating} or consider rule-based approaches \citep{pho2015distractor}. Following standard reference-based evaluation, n-gram overlap metrics such as BLEU \citep{papineni-etal-2002-bleu}, ROUGE \citep{Lin2004ROUGEAP} and METEOR \citep{banerjee-lavie-2005-meteor} have been considered, where these metrics measure the overlap between generated distractors and the distractors from a set of human-annotated ground truth sequences. However, having reference-based distractor evaluation approaches has notable shortcomings \citep{moon2022evaluating}. In particular, for a given multiple-choice question, the set of annotated distractors is unlikely to span the set of all possible good distractors, and some options may get unfairly penalised simply because no similar ones exist in the annotated set. 

We therefore focus on decomposing the distractors in terms of individual qualities (incorrectness, plausibility and diversity) and consider quality as an amalgamation of the above. These approaches have been considered for the assessment of alternate qualities of questions. \citet{dugan2022} investigate answer-agnostic generated questions in terms of the qualities of relevance, interpretability and acceptability with comparison against human markers. \citet{raina2022multiple} assess multiple-choice questions in terms of grammatical fluidity, answerability, diversity and complexity. Our work specifically explores distractor assessment in multiple-choice questions with a focus on automated assessment.

%% file: background.tex
\section{Multiple-Choice Comprehension}


In this section, we describe the multiple-choice reading comprehension task, and the architecture of standard machine reading comprehension systems. Note that the machine reading comprehension system will later be leveraged in several components of the DAF (see Section \ref{sec:dist_assess}).

\subsection{Multiple-choice comprehension task}
\label{sec:mrc}
Multiple-choice reading comprehension is a common examination format that aims to measure the reading comprehension abilities of candidates. Given question $Q$ and passage of textual information, context $C$, candidates have to select the correct answer from a discrete set of options $\{O\}$. The correct answer $y_{ans}$ is then the option where the information in the passage is consistent with the question.

\subsection{Machine reading comprehension}


\begin{figure*}[t!]
    \centering
    \includegraphics[width=\linewidth]{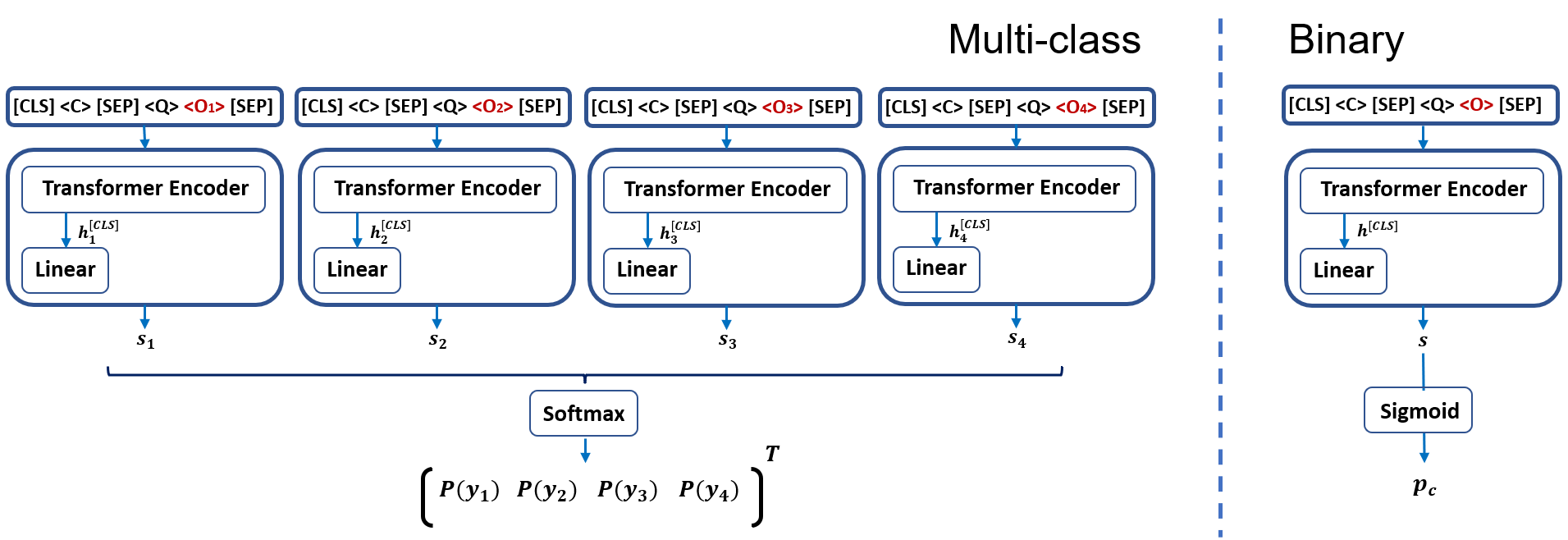}
    \caption{Model architectures of multi-class and binary multiple-choice machine reading comprehension systems with context, $C$, question, $Q$ and options, $\{O\}$.}
    \label{fig:model_arch}
\end{figure*}

Machine reading comprehension (MRC) refers to building automatic systems for performing the reading comprehension task. For multiple-choice reading comprehension, state-of-the-art machine reading systems \citep{Zhang2021RetrospectiveRF, Yamada2020LUKEDC, Zaheer2020BigBT, wang2021logicdriven} have demonstrated human-level performance on public benchmarks \citep{Clark2018ThinkYH, Lai2017RACELR, Trischler2017NewsQAAM, Yang2018HotpotQAAD}. In this work, we consider two variations of the approach:
\newline\newline\noindent 
\textbf{Multi-class MRC:} A standard approach for machine reading comprehension \citep{Yu2020ReClorAR, raina-gales-2022-answer} is to predict a probability distribution over the options, as shown in Figure \ref{fig:model_arch}. For this method, the context, question and a particular option are concatenated together and fed through a standard transformer encoder \citep{vaswani2017attention}. The hidden representation output by the transformer encoder is then passed through a linear layer to return a scalar score. This process is repeated for each option in turn, and a softmax function then returns a discrete probability distribution over the answer options. Note that the weights are shared across each of the four versions of the transformer encoder and linear layers. During inference, the answer option with the largest probability mass is selected as the system prediction.
\newline\newline\noindent 
\textbf{Binary MRC:} As an alternative to the multi-class approach, we also consider a binary multiple-choice machine reading comprehension system, as suggested by \citet{ghosal2022two}. This approach is similar to the multi-class approach, however instead of the softmax at the final stage, the binary approach applies a sigmoid to the logit scalar score. This approach therefore determines whether a given option is the correct answer, instead of determining which of the options is the correct one (which the multi-class approach does). At inference time, the answer option with the greatest probability is selected as the output from the model.
\newline\newline\noindent 
Both the binary MRC and the multi-class MRC systems have identical fundamental structures, however differ in how the options are normalised before the final probability output. The multi-class approach leverages the fact that only one of the options is correct, and so the probability of the four options are normalised relative to each other. The binary MRC system does not. Therefore for tasks that are objective in nature and depend on the single option only, such as incorrectness detection, we leverage the binary MRC system. However, the multi-class MRC system is better suited for tasks where we want to assess options relative to each other and the correct answer, such as plausibility assessment. Note, the binary MRC system can be built from a single unit of the multi-class MRC system.


\section{Distractor Assessment Framework}
\label{sec:dist_assess}


As discussed in Section \ref{sec:related}, n-gram reference-based assessment metrics may not be appropriate for distractor assessment as 1) the number of valid distractors for a given multiple choice reading question is vast, which a limited set of references may fail to capture; 2) they are only valid when a set of reference distractors are available, which requires human intervention and limits the advantages of automatic distractor evaluation. The next section discusses the DAF, which uses three different reference-free methods to estimate the quality of distractors' incorrectness, plausibility and diversity independently. 
\newline\newline 
\noindent \textbf{Incorrectness}
For multiple-choice reading comprehension questions, distractors must by definition be incorrect, and the answer option must be the only valid answer in the set of options. For automated multiple-choice question generation pipelines, it is particularly important to ensure that any generated distractors satisfy the requirement of being incorrect and do not cause a subjective interpretation of the question. Here, we detail the approach used to assess whether distractors satisfy the incorrectness requirement.

The binary MRC system from Section \ref{sec:mrc} is used to assess incorrectness. The system returns a probability score, $p_c$, which is the probability that the system thinks the given option is correct. An appropriate threshold, $\tau$ can then be selected, such if the probability score is less than the threshold, a distractor is deemed to satisfy the incorrectness requirement, as indicated in Equation \ref{eq:incorrect} (where $y\in\{\texttt{incorrect, correct}\}$ denotes the binary output decision).

\begin{equation}
\small
\label{eq:incorrect}
    y = \begin{cases}
        \texttt{incorrect}, & \text{if } p_c<\tau \\
        \texttt{correct}, & \text{otherwise}
    \end{cases}
\end{equation}

The selected threshold is a design choice of the test creator depending on how stringent the incorrectness criteria should be. For example, the incorrectness detector can be the first stage of a test creation pipeline to filter out the generated questions with multiple options that could be valid. The test creator may then select an operating point with low precision and high recall (in terms of incorrectness) in order to capture a larger pool of questions which should be considered in the subsequent stages of the evaluation pipeline. Conversely, for high-stakes educational settings, an operating point which leads to higher precision at the cost of recall may be preferred.
\newline\newline


\noindent \textbf{Plausibility} As emphasised by \citet{qiu2020automatic}, good quality distractors should be both incorrect yet also plausible and not obviously invalid. Unlike the binary incorrectness metric, plausibility is a continuous property, and distractors can be plausible to different degrees. For assessing the plausibility of a distractor within a distractor set, one can consider the model confidence of the multi-class multiple-choice machine reading comprehension system. The motivation for this approach is that a high confidence score is one that the MRC system finds more plausible, which one can assume would be similar for real candidates. We further define the plausibility score as the sum of confidence scores corresponding to each of the distractors in a question, which can similarly be calculated as the difference between 1 and the confidence score associated with the correct answer. This is expressed in Equation \ref{eq:plaus}, where $P_{\theta}$ denotes the probability distribution learnt by the multi-class machine reading comprehension system.


\begin{equation}
\small
\label{eq:plaus}
    \texttt{plausibility} = 1 - \max_{y} P_{\theta}\left(y|C,Q,\{O\}\right)
\end{equation}


\noindent \textbf{Diversity} As a human candidate, when attempting a multiple-choice question all four options are considered together. If distractors are similar or identical, then one can eliminate multiple options simultaneously using the same information, limiting the amount of comprehension that a question may require. Therefore, it becomes increasingly important to ensure the distractors are diverse. Particularly, diversity has been demonstrated to be a concern for automated question generation systems \citep{raina2022multiple}, where systems are quite susceptible to frequently generating repeated distractors. This demands a need for automated approaches to determine the diversity amongst distractors to select, or at least be aware of, the distractor set with the maximum diversity.

In this work, the BERT Equivalence Metric (BEM) \citep{bulian2022tomayto} is leveraged for assessing the diversity of the distractors. BEM is a semantic similarity measure for question answering, where the equivalence score between an answer and the reference is returned. BEM takes the text of a predicted answer, the text of the answer option and the question, concatenates them together and a BERT system then returns a scalar score, $0\leq e\leq 1$. This score captures the equivalence between the candidate and the reference, where a score of $e=1$ indicates the candidate and the reference are identical while $e=0$ indicates the candidate and the reference are completely semantically different. BEM is trained explicitly on an answer equivalence dataset and has been shown to out-perform zero-shot equivalence measures such as the BERTScore from \citet{zhangbertscore}. 

In this work, BEM is applied pair-wise to all possible pairs of distractors in a given question. The context is not concatenated to the question since initial experiments demonstrated that the long contexts diluted the differences between pairs of distractors. Since BEM is not order invariant, we average the output from BEM with both orderings for the pair of distractors considered. The overall diversity is quoted as the 1 minus the average pairwise BEM scores between the distractors, as indicated by Equation \ref{eq:diversity} where the $K$ distractors associated with a given question are denoted as $\{d_1, d_2, \hdots, d_K\}$.
\begin{equation}
\small
\label{eq:diversity}
    \texttt{diversity} = 1. - \sum_{i=1}^{K}\sum_{j=1,j\neq i}^K \frac{\texttt{BEM}[d_i,d_j,Q]}{K^2-K}
\end{equation}

%% file: experiments.tex
\section{Experiments}
\subsection{Data}
\label{sec:data}
\noindent \textbf{RACE++}: RACE++ \citep{pmlr-v101-liang19a} is a large-scale machine reading comprehension dataset of real questions used in middle school (RACE-M), high school (RACE-H) and college level (RACE-C). There are 4 options per question with a single option as the correct for each. Table \ref{tab:race++_split} details the train, validation and test splits used for training and testing of the multiple-choice reading comprehension datasets. 

\begin{table}[h]
\small
    \centering
    \begin{tabular}{c|rrr}
        \toprule
        subset & train & valid & test \\
        \midrule
        RACE-M & 25,241  & 1,436 & 1,436 \\
        RACE-H & 62,445  & 3,451 & 3,498 \\
        RACE-C & 12,702  & 712   & 708 \\
        \midrule
        RACE++ & 100,388 & 5,599 & 5,642 \\
        \bottomrule
    \end{tabular}
    \caption{Data splits for RACE++. RACE++ is composed of questions at the middle school (M), high school (H), and college (C) level.}
    \label{tab:race++_split}
\end{table}

\noindent \textbf{CMCQRD}\footnote{\url{https://www.englishlanguageitutoring.com/}.}: The Cambridge Multiple-Choice Questions Reading Dataset (CMCQRD) \citep{CMCQRD-2023} is a small-scale multiple-choice reading comprehension evaluation dataset from the pre-testing stage partitioned into grade levels B1 to C2 on the Common European Framework of Reference for Languages (CEFR). Additionally, a subset of the CMCQRD dataset has candidate distributions available. We perform our experiments only on this subset of questions as analyzed in \citet{liusie2023analysis}. The statistics of these questions are given in Table \ref{tab:camchoice}. 

\begin{table}[h]
\small
    \centering
    \begin{tabular}{c|rr}
        \toprule
        subset & contexts & questions \\
        \midrule
        B1 & 23 & 115 \\
        B2 & 37 & 222 \\
        C1 & 12 & 72 \\
        C2 & 6 & 39 \\
        \midrule
        CMCQRD & 78 & 448 \\
        \bottomrule
    \end{tabular}
    \caption{Splits of CMCQRD subset (with candidate distribution) of data between CEFR levels.}
    \label{tab:camchoice}
\end{table}

\subsection{Training}

For multi-class MRC, we take the ELECTRA pretrained language model \citep{clark2020electra} (specifically \texttt{ELECTRA-large} \footnote{Available at: \url{https://huggingface.co/google/electra-large-discriminator}}) and train the system with cross-entropy loss on the train split of RACE++, with the best epoch selected using the RACE++ validation split. Following \citet{raina-gales-2022-answer}, the model is trained using the AdamW optimizer, a batch size of 4, learning rate of 2e-6 and a maximum of 3 training epochs. All inputs are truncated to 512 tokens, and all processing is performed on NVIDIA V100 graphical processing units. We consider ensembles of 3 models for each system. For the binary MRC system, a single unit of the trained multi-class MRC system is used with the softmax layer removed and a sigmoid at the output instead (mimics Figure \ref{fig:model_arch}). \footnote{Initial experiments trained a separate system for binary MRC where each option was reformatted as individual data points with either a label of correct (answer) or incorrect (distractor). However, this system generalized poorly to CMCQRD despite good performance on the RACE++ dataset.}

\section{Results}
In this section, we present results for assessing incorrectness, plausibility and diversity as part of the DAF for standard multiple-choice reading comprehension datasets.


Table \ref{tab:baseMCRC} presents the baseline performance of the MRC system on the RACE++ and CMCQRD test sets. Overall, the MRC system ports across well from RACE++ to CMCQRD, getting an accuracy of 85\% on RACE++ and 74\% on CMCQRD. It is also apparent that for both datasets, the accuracy of the MRC system degrades for more challenging questions by approximately 7\% from RACE-M to RACE-C and 25\% from CEFR level B1 to C2.

Table \ref{tab:baseMCRC} further presents the newly proposed incorrectness, plausibility and diversity scores using the described approaches applied to both multiple-choice reading comprehension datasets. For each question in each dataset, the distractors for the question are considered to be the set of `generated' distractors (first stage of distractor generation in the pipeline of Figure \ref{fig:daf_pipeline}) for which the incorrectness, plausibility and diversity scores need to be calculated. For incorrectness, each distractor is classified as either incorrect or correct based on the optimal operating point threshold of performance (see Table \ref{tab:corr_detector}) which is a value of $\tau=0.25$ for RACE++ and $\tau=0.04$ for CMCQRD. Hence, the overall incorrectness score is the percentage of distractors that are categorized as incorrect (higher is better). Plausibility (Equation \ref{eq:plaus}) and diversity (Equation \ref{eq:diversity}) scores are averaged across all the questions in the dataset.

\begin{table}[htbp!]
\small
\centering
\begin{tabular}{l|c|ccc}
\toprule
Dataset & Acc. & Incorr. & Plaus. & Divers. \\
\midrule
RACE++ & 85.0  & 91.8 & 15.0 & 74.1 \\
RACE-M & 88.1  & 93.8 & 11.8 & 66.8 \\
RACE-H & 84.4  & 91.0 & 15.7 & 75.7 \\
RACE-C & 81.6  & 91.7 & 18.0 & 81.0 \\
\midrule
CMCQRD & 74.3  & 86.7 & 27.7 & 78.2 \\
B1 & 90.4  & 85.5 & 11.9 & 75.7 \\
B2 & 73.4  & 86.9 & 30.0 & 78.0 \\
C1 & 56.9  & 87.5 & 40.9 & 80.3 \\
C2 & 64.1  & 87.2 & 37.0 & 82.8 \\
   \bottomrule
    \end{tabular}
\caption{Ported accuracy of the MRC system trained on RACE++. For proposed distractors (in the dataset), incorrectness rate, average plausibility and diversity scores are reported as percentages.}
\label{tab:baseMCRC}
\end{table}

It is observed that the incorrectness rate remains consistent across all the splits for RACE++. A similar consistency is evident on the CMCQRD dataset. In general, it can be seen that the plausibility scores tend to be higher for more challenging questions for both RACE++ and CMCQRD. This is potentially explainable by the fact that more challenging questions can expect to have a greater probability mass attributed to the distractors compared to the correct answer option. Loosely, the average diversity score follows a similar pattern where more challenging questions can expect to have more diverse distractors. Possibly a low diversity in the distractors offers fewer opportunities to \textit{distract} the candidates.

We have presented the incorrectness, plausibility and diversity scores on the RACE++ and CMCQRD datasets. The subsequent sections aim to provide a form of verification for each of these metrics to demonstrate they are suitable for the respective qualities that they are assessing.

\subsection{Assessing correctness detector}

This section assesses the accuracy of the correctness detector which is used for measuring the incorrectness rate. To assess the accuracy, we assume that the allocation of answer options as either distractors or the correct answer are the ground-truth binary labels. Table \ref{tab:corr_detector} assesses how well the correctness detector performs on RACE++ and CMCQRD datasets using the optimal F1 score for this binary classification task.

\begin{table}[htbp!]
\centering
\small
\begin{tabular}{l|ccc}
\toprule
& Precision & Recall & F1 \\
\midrule
RACE++ & 80.1 & 72.7 & 76.2 \\
CMCQRD & 62.2 & 65.8 & 64.0 \\
   \bottomrule
    \end{tabular}
\caption{Performance for the correctness detector.}
\label{tab:corr_detector}
\end{table}

\begin{figure}[h]
    \centering
    \includegraphics[width=0.8\columnwidth]{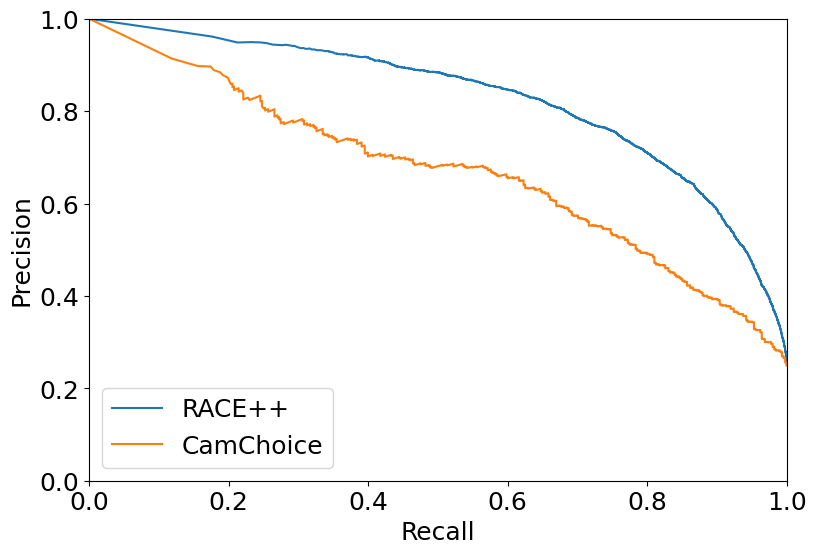}
    \caption{Precision-Recall curve for correctness detector on RACE++ and CMCQRD.}
    \label{fig:pr}
\end{figure}

Figure \ref{fig:pr} presents the precision-recall curve of the correctness detector on both the RACE++ and CMCQRD datasets. From both Figure \ref{fig:pr} and Table \ref{tab:corr_detector}, the performance of the correctness detector is sensible, with performance on CMCQRD lagging RACE++ demonstrated by top F1 scores of 76\% and 64\% on RACE++ and CMCQRD respectively. In line with these single-value summaries, the CMCQRD precision-recall curve undercuts the RACE++ precision-recall curve for all recall rates.  

\begin{figure}[htbp!]
     \centering
     \begin{subfigure}[b]{0.49\columnwidth}
         \centering
         \includegraphics[width=\columnwidth]{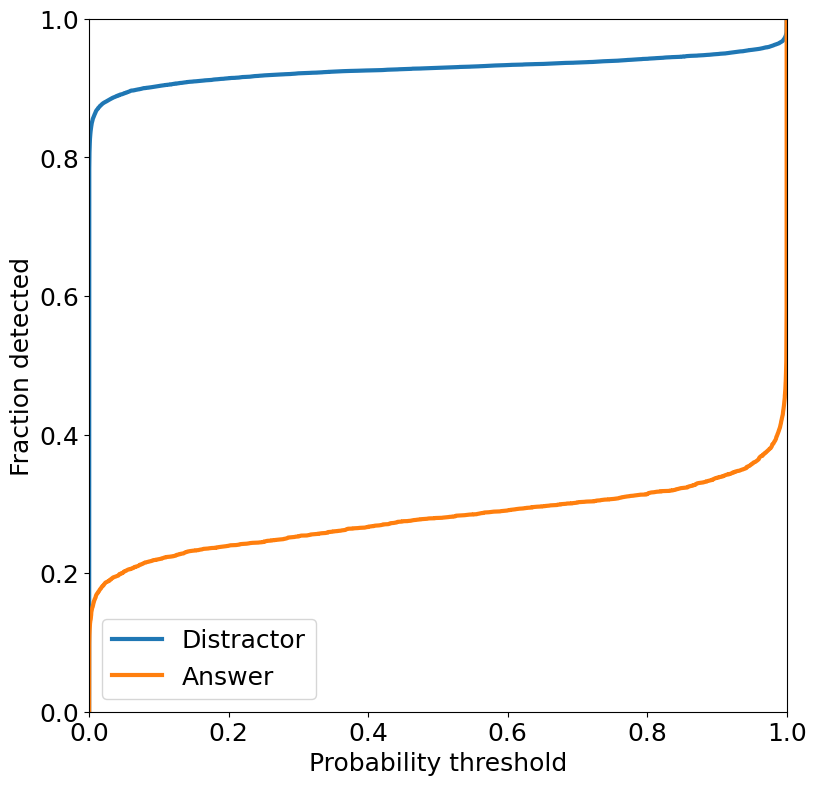}
         \caption{RACE++.}
         \label{fig:gde}
     \end{subfigure}
     \hfill
     \begin{subfigure}[b]{0.49\columnwidth}
         \centering
         \includegraphics[width=\columnwidth]{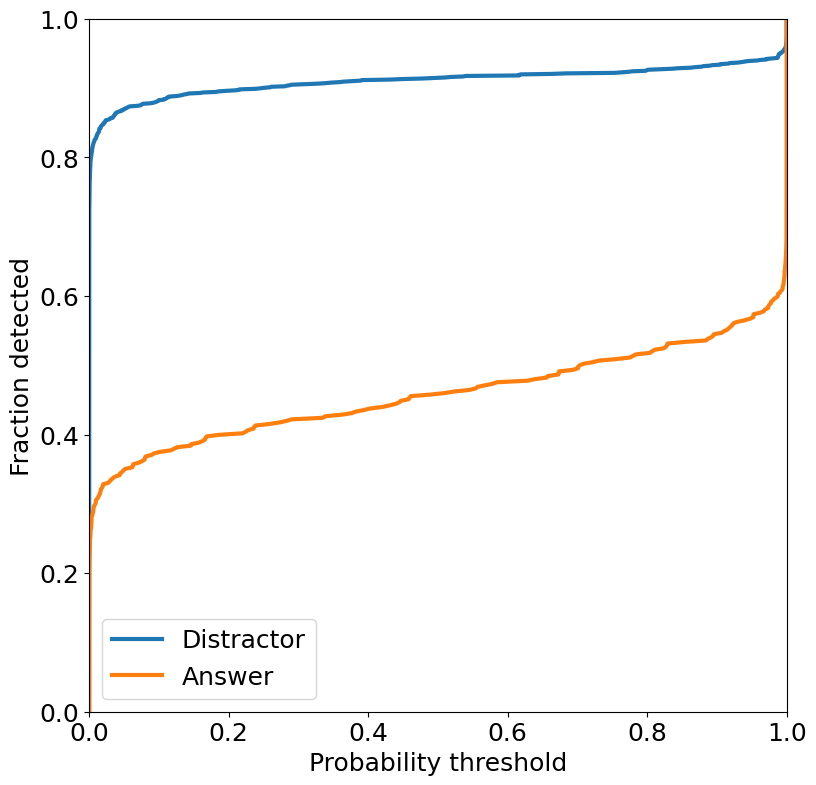}
         \caption{CMCQRD.}
         \label{fig:gade}
     \end{subfigure}
        \caption{Operating chart for correctness detector on RACE++ and CMCQRD.}
        \label{fig:opChart}
\end{figure}

Figure \ref{fig:opChart} further presents an operating chart for the correctness detector. The chart sweeps the threshold for the binary MRC system from 0 to 1 and identifies the fraction of distractors and answer options that are captured cumulatively. As expected, for both the RACE++ and CMCQRD operating charts, the `distractor' curve significantly leads the `answer' curve. The operating charts offers content creators a means to choose an operating threshold; a low threshold on correctness may guarantee that only real distractors are captured but also reduces the pool of distractors that are considered in the review process.

\subsection{Verification of plausibility/diversity via ChatGPT}

Recently, generative large-scale foundation models \citep{brown2020language, chowdhery2022palm, scao2022bloom}, such as the popularized ChatGPT, have demonstrated state-of-the-art performance across a large range of natural language tasks in zero-shot and few-shot settings. These models are particularly impressive at successfully completing tasks that they have never seen before.

\begin{figure}[t!]
    \centering
    \includegraphics[width=1.0\columnwidth]{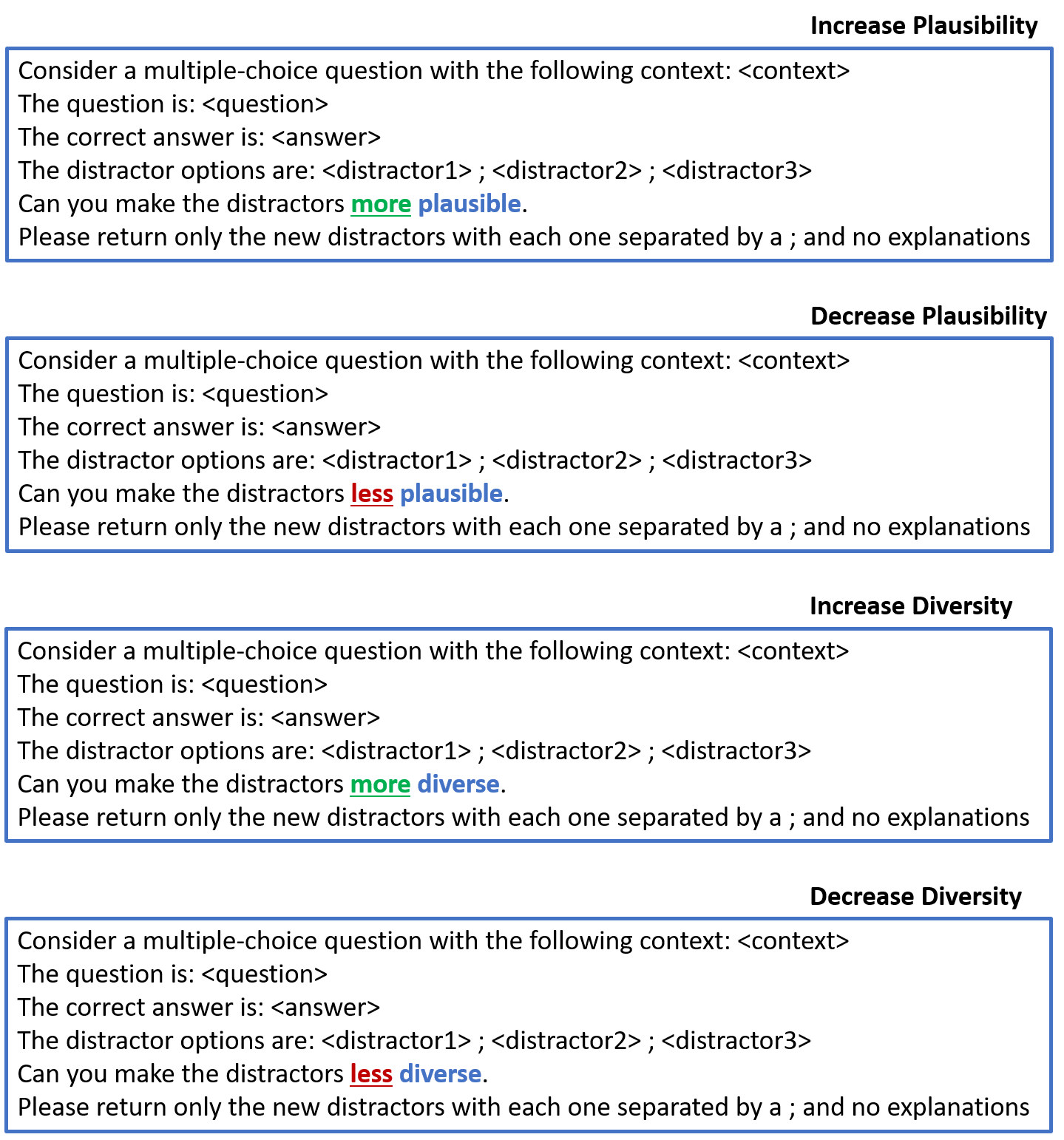}
    \caption{Prompting ChatGPT for more/less plausible/diverse distractors.}
    \label{fig:gpt_prompt}
\end{figure}

However, there remain several practical challenges in using foundation models such as ChatGPT.
1. There are concerns with security of the data as confidential data cannot be taken off-site. This is an important consideration in the educational setting as often the trial questions to be assessed will form a core component of a live standardized test. 2. Access via an API (Application Program Interface) means the input/output is restricted as well as the risk of evolving with limited warning, jeopardizing an exterior infrastructure developed to interact with the API. 3. There is a continual cost of interacting with API access which may limit the scalability in deployment. 4. Zero-shot performance can be challenging to tune to specific tasks(s) of interest.
Therefore, we do not employ ChatGPT as a direct assessment approach for the DAF. It is necessary for any automated assessment approach to be local so that there is complete control over the model. Instead, ChatGPT is considered here as a validation process for the DAF that in itself bypasses ChatGPT's challenges.

Here ChatGPT, specifically \texttt{gpt-3.5-turbo} \footnote{\url{https://platform.openai.com/docs/models/gpt-3-5}}, is employed as an approach to verify the proposed plausibility and diversity assessment metrics. ChatGPT is given a standard multiple-choice reading comprehension question. The foundation model is then requested to refine the choice of the distractors to make them more/less plausible or diverse. Figure \ref{fig:gpt_prompt} presents the prompts.

By probing ChatGPT to create alternatives for the distractors, it is useful to check the agreement of ChatGPT's interpretation of plausibility and diversity with the proposed assessment metrics.

\begin{table}[htbp!]
\small
\centering
\begin{tabular}{l|cccc}
\toprule
System  & All & M & H & C \\
\midrule
Vanilla & 85.0 & 88.1 & 84.4 & 81.6 \\
\midrule
Increase plausibility & 74.6 & 77.5 & 73.6 & 73.7 \\
Decrease plausibility & 84.0 & 85.3 & 83.3 & 85.2 \\
\midrule
Increase diversity & 74.6 & 78.5 & 73.7 & 71.3 \\
Decrease diversity & 62.0 & 68.3 & 59.7 & 60.9\\
   \bottomrule
    \end{tabular}
\caption{Accuracy of ensemble on test split of RACE++ (RACE-M, RACE-H, RACE-C) using the multi-MRC system after probing ChatGPT to refine the distractors in terms of plausibility and diversity.}
\label{tab:probechatgpt}
\end{table}

From Table \ref{tab:probechatgpt}, the accuracy of RACE++ trained system is impacted by exchanging the distractors with variants provided by ChatGPT. Prompting ChatGPT to generate more plausible distractors leads to the accuracy of the MRC system dropping by up to 10\% as the altered questions on average are more challenging. In contrast, prompting ChatGPT to decrease the plausibility has less of an impact on the behaviour of the MRC system's accuracy. By prompting ChatGPT to increase or decrease the diversity of the distractors, there is an observed drop in the MRC system accuracy, particularly for less diverse distractors of more than 20\%. 

\begin{table}[htbp!]
\small
\centering
\begin{tabular}{l|cc}
\toprule
System  & Standard & Context-free \\
\midrule
Vanilla & 85.0 & 57.0 \\
\midrule
Increase plausibility & 74.6 & 39.3 \\
Decrease plausibility & 84.0 & 54.4 \\
\midrule
Increase diversity & 74.6 & 42.3 \\
Decrease diversity & 62.0 & 38.7 \\
   \bottomrule
    \end{tabular}
\caption{Impact of world knowledge after probing ChatGPT to refine the distractors in terms of plausibility and diversity on the RACE++ test set.}
\label{tab:worldknowledge}
\end{table}

In Table \ref{tab:worldknowledge}, the impact of world knowledge in reading comprehension \citep{liusie-etal-2023-world} is explored for the ChatGPT generated distractors. Here, a context-free system (no access to the context) measures to what extent a question relies on using knowledge outside the context to determine the correct answer. With all values substantially above the random performance of 25\%, for both the original questions and the probed version of the questions, there is significant scope to leverage world knowledge to answer the questions. 

\begin{figure}[h]
    \centering
    \begin{subfigure}[t]{0.48\columnwidth}
        \centering
    \includegraphics[width=1.0\columnwidth]{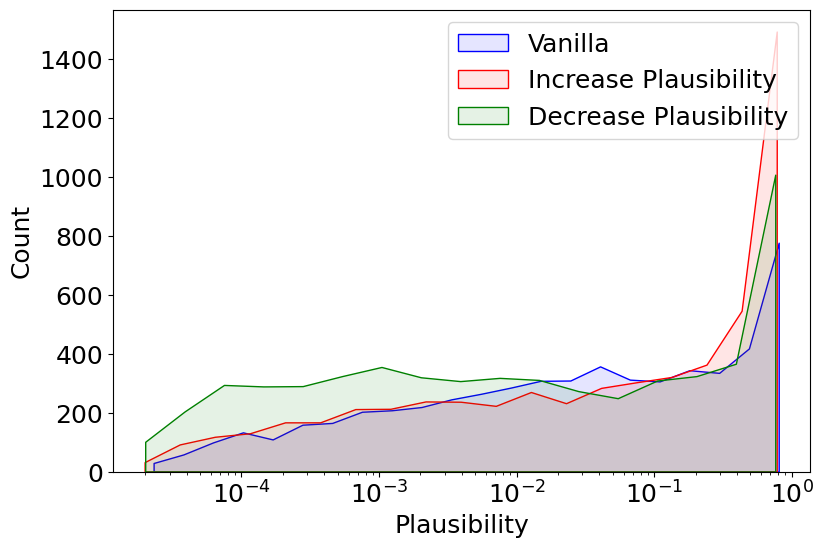}
    \caption{Plausibility.}
    \end{subfigure}
    ~
    \begin{subfigure}[t]{0.48\columnwidth}
        \centering
\includegraphics[width=1.0\columnwidth]{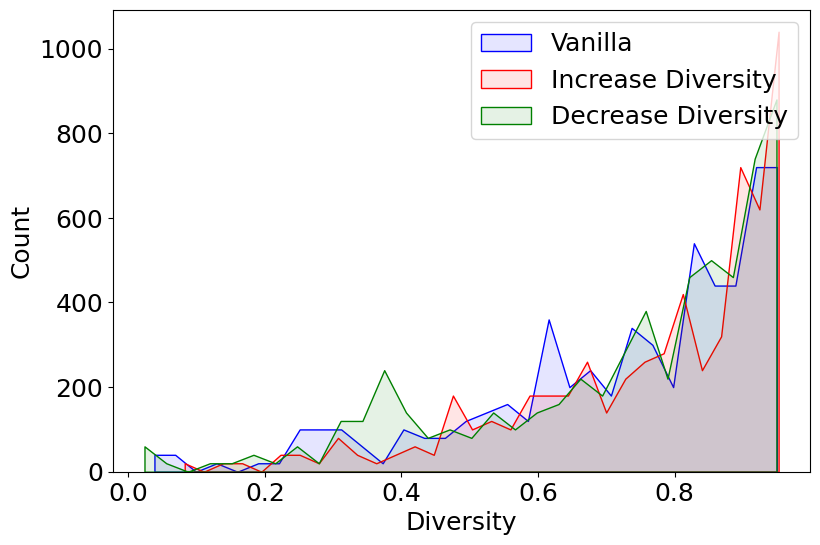}
    \caption{Diversity.}
    \end{subfigure}
    \caption{Impact on distribution of plausibility and diversity by probing ChatGPT on the RACE++ dataset. See Equations \ref{eq:plaus} and \ref{eq:diversity} for the definitions of plausibility and diversity respectively.}
    \label{fig:probe_chatgpt_dist}
\end{figure}

The distribution of plausibilities and diversities in Figure \ref{fig:probe_chatgpt_dist} further demonstrates that there is an upward shift with the increased plausibility/diversity variants of the distractors.

\begin{table}[htbp!]
\small
\centering
\begin{tabular}{ll|rr}
\toprule
setA & setB & $>$ plaus. & $>$ div.  \\
\midrule
Increase & Vanilla & 57.6 & 60.7 \\
Vanilla & Decrease & 63.2 & 53.2 \\
Increase & Decrease & 69.1 & 63.3 \\
   \bottomrule
    \end{tabular}
\caption{Fraction of examples for which plausibility and diversity of distractors in a question  for setA is $>$ setB.}
\label{tab:comp}
\end{table}

Finally, Table \ref{tab:comp} presents the impact of refining the distractors at an individual question level. For example, the increased plausibility versions of the distractors compared to the decreased plausibility versions have a high plausibility score for 69\% of the RACE++ questions. It seems it is challenging to be able to increase plausibility and decrease diversity, while it is relatively easier to decrease plausibility and increase diversity. With all scores above 50\%, it suggests that there is alignment between ChatGPT's interpretation of plausibility/diversity and the assessment approaches for these qualities.

\subsection{Verification of plausibility via candidate distribution}

As in Section \ref{sec:dist_assess}, the plausibility of distractors is assessed using the probability confidence scores distribution output from a multiple-choice machine reading comprehension system. The claim is that a higher confidence score suggests that a distractor is more plausible. In a practical sense, the plausibility scores for the distractors should correspond with how likely a candidate taking a test is to select the distractors.
CMCQRD (see Section \ref{sec:data}) includes candidate distributions over multiple-choice questions. Therefore, the human candidate distributions are used to verify whether the plausibility scores from a standard multiple-choice reading comprehension system correspond with candidates' interpretation of the plausibility of distractors.

We consider two comparison methods for validating the plausibility scores. Intra-question: compare the ranking of distractors by system confidence and human confidence for each question. Inter-question: compare the ranking across questions of distractor confidence (see Equation \ref{eq:plaus}) by the system and the candidates.

The intra-question verification informs whether the individual distractor plausibility scores by the system can be used to identify which distractors are more convincing while the inter-question verification informs whether the system's distractor confidence is a universal measure of how convincing the distractors are for a question as a collective.

For intra-question rankings, the averaged (across questions) Spearman's rank correlation coefficients between the candidate probabilities for a set of distractors per question and the system's probabilities for the same set of distractors is 0.25. For the inter-question case, the global Spearman's rank correlation between candidate plausibility (sum of individual distractor confidences) and system plausibility is 0.22. Despite not being strong correlations (potentially due to human noise from learners taking the test), the positive values indicate that human understanding of distractor plausibility is somewhat aligned with the system's understanding.

%% file: conclusions.tex
\section{Conclusions}
This work proposes the distractor assessment framework, an automatic approach for assessing the quality of distractors on three key properties: incorrectness, plausibility and diversity. By leveraging multi-class and binary machine reading comprehension systems, and semantic similarity metrics, we propose intuitive methods for calculating automatic scores for the 3 properties. We validate the metrics by refining distractors with ChatGPT. Further there is a positive correlation indicated between candidate and system plausibilities.

\section{Acknowledgements}
This research is partially funded by the EPSRC (The Engineering
and Physical Sciences Research Council)
Doctoral Training Partnership (DTP) PhD studentship
and supported by Cambridge University Press \& Assessment (CUP\&A), a
department of The Chancellor, Masters, and Scholars
of the University of Cambridge.

%% file: appendix.tex

\section{Limitations}

This work explores automated approaches to assess the incorrectness, plausibility and diversity of distractors encompassed in a DAF. A limitation is that the current research focuses specifically on the RACE++ and CMCQRD datasets. Further work should investigate the applicability of the DAF for other multiple-choice datasets.
Second, the proposed assessment methods are verified using both candidate distributions and agreement with ChatGPT's interpretation of the qualities. Explicit at-scale human evaluation may help provide further evidence for the validity of the assessment approaches.
Finally, a more extensive and rigorous approach could have been taken to determine optimal prompts for increasing/decreasing the plausibility or the diversity of the distractors.


\section{Ethics Statement}
There are no ethical concerns with this work.